\documentclass[runningheads]{llncs}

\usepackage{float}
\usepackage{amsmath}
\usepackage{amsfonts}
\usepackage{multirow}
\usepackage[T1]{fontenc}

\usepackage{xltxtra}
\newfontfamily{\GjFont}{gj.ttf}
\newfontfamily{\MrFont}{NotoSerifDevanagari.ttf}

\usepackage{booktabs}
\usepackage{graphicx}
\usepackage{subcaption}

\newcommand{\myfirstpara}[1]{\noindent \textbf{#1.}}
\newcommand{\mypara}[1]{\vspace{0.2em} \myfirstpara{#1}}

\usepackage{hyperref}
\usepackage{color}

\begin{document}

\title{What Can Low Resource Languages \\ Learn From Each Other?}

\author{Achyuth P$^*$ \and
Kahaan Shah$^*$ \and
Chetan Arora}
\authorrunning{Achyuth P. et al.}
\titlerunning{What Can Low Resource Languages Learn From Each Other?}

\institute{
IIT Delhi, New Delhi, India\\
\email{\{achyuth.cstaff, kahaan.cstaff, chetan\}@cse.iitd.ac.in}}

\maketitle
\def\thefootnote{*}\footnotetext{These authors contributed equally to this work}

\begin{abstract}
	Despite the rapid advancement of Vision-Language Models (VLMs), their linguistic reach remains largely confined to high-resource languages, leaving the majority of the world's 7,000+ living languages on the wrong side of a growing digital divide. This disparity is especially pronounced in Optical Character Recognition (OCR), where low-resource scripts lack the massive datasets required for traditional scaling laws. We investigate OCR adaptation in extreme data-scarce regimes ($<$10K real and $<$250K synthetic images), demonstrating that conventional fine-tuning strategies often reach a performance ceiling. Our key finding reveals a structural inefficiency in language-specific adaptation: while higher layers of specialized models diverge to capture unique script nuances, the lower layers learn redundant, highly similar features. Motivated by this observation, we propose PSMC (Pre-train, Specialize, Merge, and Co-train), a data-efficient framework that capitalizes on a cross-script ``transfer effect''. Our approach first derives language-specific experts from a high-resource base model, then employs task arithmetic to fuse these experts into a unified, high-performance multilingual backbone. Extensive evaluation across 10 Indian scripts (supporting 20+ languages) shows that PSMC achieves a $\sim$2\% average improvement in Word Recognition Rate (WRR) over individual specialist models without increasing parameter count. Our results indicate that joint training in the merged latent space facilitates a constructive knowledge transfer that benefits all constituent scripts, providing a scalable pathway for inclusive VLM development. Source code and datasets will be released post-publication.
\end{abstract}

\section{Introduction} 

The recent surge in Vision-Language Models (VLMs) has enabled unprecedented capabilities in cross-modal understanding. However, these advancements have largely been asymmetric. While high-resource languages from the Global North enjoy robust support, a significant majority of the world's 7,000+ living languages remain sidelined, resulting in a widening digital divide. This disparity is particularly critical in Optical Character Recognition (OCR), which serves as the fundamental gateway for digitizing and preserving the cultural and administrative heritage of low-resource communities.

Current paradigms for extending OCR to new scripts rely on massive data scaling; a luxury not afforded by the ``long-tail'' of languages from the Global South. In extreme data-scarce regimes, where real training samples are often capped at fewer than 10,000 images, standard adaptation strategies begin to falter. Language-specific fine-tuning or delta training, while effective in high-resource contexts, often results in overfitting or catastrophic forgetting of the general visual features learned during the model's base pre-training. Consequently, these models reach a performance ceiling that fails to capture the intricate nuances of diverse scripts.

\begin{figure}[t]
	\centering
	\includegraphics[width=0.80\linewidth]{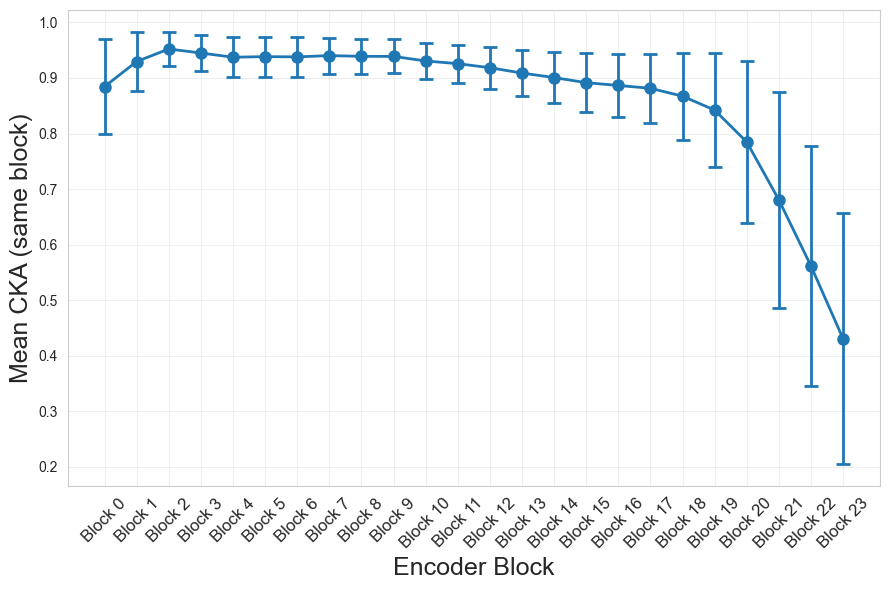}
	\caption{
		To quantitatively assess the similarity of learned representations across language-specific models, we employ Centered Kernel Alignment (CKA)~\cite{CKA}, a representation similarity metric that is invariant to orthogonal transformations and isotropic scaling. We consider a base model trained on the full English dataset and further finetuned on 250K samples (10K real, 240K synthetic) of each script. We choose 25 scene and 25 printed text images per language and pass them through each model, extracting features at each encoder layer. We then perform a CKA analysis at each layer (more details in the experiments section). The plot shows the average similarity between layers at the same index, averaged across all images and model pairs. Observe that the representations in the lower layers are very similar, while language specific features are learned in the deeper layers of each encoder. This motivates us to co-train across languages, allowing the encoder to jointly learn the early script-agnostic robust representations from varied samples while accommodating script-specific features in the final encoder layers.
	}
	\label{fig:layer-sim}
\end{figure}

Through an architectural analysis of language-specific ``specialists'', we identify a crucial structural redundancy: while the higher layers of these models diverge to master script-specific characters, the lower layers learn remarkably similar, domain-agnostic visual features (see Fig. \ref{fig:layer-sim}). This observation suggests that maintaining separate, isolated models for each language is not only parameter-inefficient but also ignores a latent opportunity for cross-linguistic synergy.

In this work, we propose PSMC (Pre-train, Specialize, Merge, and Co-train), a novel, data-efficient framework that leverages joint training across diverse scripts to overcome the limitations of data scarcity. Our approach moves beyond simple fine-tuning by first deriving language-specific experts (`S' stage) from a common pretrained base model (`P' stage), and then employing task arithmetic to fuse their weight offsets into a single, unified multilingual backbone (`M' stage). This merging process effectively ``pools'' the commonalities of different scripts while preserving their unique diagnostic features. Finally, we perform a joint co-training (`C' stage) with all the languages to disentangle representations across languages. 

By evaluating PSMC across 10 Indian scripts, covering more than 20 languages, we demonstrate that our unified model consistently outperforms individual specialist models. Our findings reveal a constructive transfer effect: by training in a merged latent space, the model leverages shared structural patterns between scripts (e.g., the shared phonetic roots and stroke styles across scripts) to improve the Word Recognition Rate (WRR) by an average of ~2\%.

\begin{table}[t]
    \centering
    \setlength{\tabcolsep}{15pt}
    \begin{tabular}{ll}
       \toprule
       \textbf{Script} & \textbf{Languages} \\
        \midrule
        \multirow{2}{*}{Devanagari} & Hindi, Marathi, Sindhi, Kashmiri, Dogri,\\ 
        & Nepali, Bodo, Maithili, Sanskrit, Konkani \\
        \midrule
        Bengali-Assamese & Bengali, Assamese, Santali, Manipuri \\
        \midrule
        Odia & Odia \\
        \midrule
        Gujarati & Gujarati \\
        \midrule
        Kannada & Kannada \\
        \midrule
        Malayalam & Malayalam \\
        \midrule
        Meitei & Manipuri \\
        \midrule
        Gurmukhi & Punjabi \\
        \midrule
        Tamil & Tamil \\
        \midrule
        Telugu & Telugu\\
        \bottomrule
    \end{tabular}
    \vspace{0.5em}
    \caption{Indic Languages and Scripts}
    \label{tab:scripts}
\end{table}

\mypara{Contributions}
Our contributions are summarized as follows:
\begin{itemize}
	\item We identify a hierarchical redundancy in specialized OCR models, providing a motivation for model merging in data-constrained environments.
	\item We introduce PSMC, a four-stage framework that utilizes task arithmetic to create a single high-performance multilingual OCR model.
	\item We provide a systematic evaluation across 10 low-resource Indian scripts, proving that joint refinement in a merged space yields superior results to isolated specialization.
	\item We release the code and datasets used in this study to facilitate the expansion of OCR support for the global long-tail of languages.
\end{itemize}

\section{Related Work}
\begin{table}[t]
\centering
	\setlength{\tabcolsep}{4pt}
    \begin{tabular}{cccc}
        \toprule
        \textbf{Dataset} & \textbf{Languages} & \textbf{Train Size} & T\textbf{est Size} \\
        \midrule
        \multirow{3}{90pt}{\centering IndicSTR-Roadside \\ (Real, Scene) \cite{Indic_Roadside}} & Bengali, Gujarati \footnotemark[1], Hindi, & \multirow{3}{20pt}{\centering175K} & \multirow{3}{20pt}{\centering50K}\\
        & Kannada, Malayalam, Marathi & & \\
        & Oriya, Punjabi, Tamil, Telugu &  &\\
        \midrule
        \multirow{4}{90pt}{\centering IndicSTR12 \\ (Real, Scene) \cite{IndicSTR12}} & Assamese, Bengali, Odia & \multirow{4}{20pt}{\centering20K} & \multirow{4}{20pt}{\centering7K}\\
        & Marathi, Hindi, Kannada & & \\
        & Urdu, Telugu, Malayalam &  &\\
        & Gujarati, Punjabi, Tamil & & \\
        \midrule
        \multirow{4}{90pt}{\centering Mozhi \\ (Real, Printed) \cite{Towards_Deployable}} & Assamese, Bengali, Odia, Manipuri & \multirow{4}{20pt}{\centering960K} & \multirow{4}{20pt}{\centering120K}\\
        & Marathi, Hindi, Kannada & & \\
        & Urdu, Telugu, Malayalam &  &\\
        & Gujarati, Punjabi, Tamil & & \\
        \midrule
        \multirow{4}{90pt}{\centering IndicSTR12-Synth \\ (Synthetic, Scene) \cite{IndicSTR12}} & Assamese, Bengali, Odia, Manipuri & \multirow{4}{20pt}{\centering24M} & \multirow{4}{20pt}{\centering6M}\\
        & Marathi, Hindi, Kannada & & \\
        & Urdu, Telugu, Malayalam &  &\\
        & Gujarati, Punjabi, Tamil & & \\
        \midrule
        \multirow{4}{90pt}{\centering IIIT-Synthetic-R \\ (Synthetic, Printed) \cite{ILOCR}} & Assamese, Bengali, Odia, Manipuri & \multirow{4}{20pt}{\centering6M} & \multirow{4}{20pt}{\centering4.5M}\\
        & Marathi, Hindi, Kannada & & \\
        & Telugu, Malayalam &  &\\
        & Gujarati, Punjabi, Tamil & & \\
        \bottomrule
    \end{tabular}
    \vspace{0.5em}
    \caption{Public Datasets Used}\label{tab:public_datasets}
\end{table}

\footnotetext[1] {We found that the Gujarati set had several Hindi samples, mislabelled GT and samples with no visible text. We present results after omitting these.}
\myfirstpara{Multimodal OCR and Transfer Learning}
Recent advancements in Optical Character Recognition (OCR) have been dominated by large-scale Vision-Language Models (VLMs) from industry, such as Qwen-VL \cite{bai2025qwen3vltechnicalreport}, InternVL \cite{wang2025internvl35advancingopensourcemultimodal}, and DeepSeek-OCR \cite{wei2025deepseekocrcontextsopticalcompression}. While these models achieve impressive results on holistic page-level understanding through massive data scaling, they diverge from the traditional OCR paradigm that prioritizes high-fidelity word- and line-level recognition - a granular focus still essential for accurate archival digitization.
Parallel to these industrial scales, research has increasingly turned toward transfer learning to address the data scarcity inherent in many scripts. In the non-Indic context, Laurent and Lauar \cite{lauar2024spanishtrocrleveragingtransfer} demonstrated the efficacy of adapting a pretrained TrOCR \cite{trocr} backbone to create specialized, single-language models for low-resource Spanish. Within the Indic domain, Baek et al. \cite{10445946} investigated multilingual training on the MLT-19 dataset using CNN-based architectures; however, while they noted broad benefits to multilingualism, their approach failed to surpass established benchmarks for Bangla and Hindi. Similarly, Gunna et al. \cite{IndicTransfer} explored transfer learning trajectories from high-resource English and Hindi to other Indic scripts using CNN and Bi-LSTM backbones on the MLT-17 and MLT-19 datasets.
We build upon these foundational efforts but depart from the traditional ``one-to-one'' transfer paradigm. Rather than merely adapting a model to a single target, we investigate how co-training and task arithmetic can synthesize knowledge from multiple language-specific experts. Our work demonstrates that by strategically merging these representations, we can develop a unified multilingual model that not only matches but exceeds the performance of isolated specialist models.

\mypara{Indic Languages OCR Datasets}
The development of robust OCR for Indic scripts has been supported by several key datasets, varying significantly in scale and domain. At the word level, IndicSTR-Roadside~\cite{Indic_Roadside} and Mozhi~\cite{Towards_Deployable} represent the most substantial resources; the former provides 17,500 scene-text samples across 10 languages, while the latter offers 80,000 printed-text samples across 12 languages. Additionally, the IndicSTR12~\cite{IndicSTR12} dataset, an extension of IIIT-ILST~\cite{IL-SCENETEXT_MINESH}, contributes 27,000 scene-text images (for 12 languages) along with a large-scale synthetic corpus of 2M training and 500K test samples per language. However, data scarcity remains a critical bottleneck for multilingual research. Established multilingual benchmarks such as MLT-17~\cite{MLT17} and MLT-19~\cite{MLT19} contain only approximately 4,000 training samples and, notably, lack publicly available annotated test sets for prominent languages like Hindi and Bengali. While recent efforts have attempted to address the ``long-tail'' of scripts, such as Mozhi-LR~\cite{mozhi-lr} for low-resource printed text and IndicVisionBench~\cite{faraz2026indicvisionbench} for page-level VLM evaluation, these resources are currently not accessible to the public.

\mypara{Indic Language OCR models}
The current landscape of word-level Indic OCR is primarily defined by language-specific ``specialists'' developed alongside major data releases. The authors of IndicSTR-Roadside~\cite{Indic_Roadside} and the Mozhi dataset~\cite{Towards_Deployable} established the prevailing state-of-the-art (SOTA) by training individual expert models using both CNN-based and Transformer-based architectures. While these models provide high-fidelity recognition for their respective target scripts, they are typically optimized in isolation, lacking a unified multilingual representation. Earlier efforts in the field focused on benchmarking using the IIIT-ILST~\cite{IL-SCENETEXT_MINESH} and IndicSTR12~\cite{IndicSTR12} datasets. For instance, Saluja et al.~\cite{OCR-on-the-go} utilized the IIIT-ILST dataset to benchmark their OCR system. These earlier works largely relied on traditional deep learning pipelines, often combining CNNs for feature extraction with RNNs for sequence modeling, before the recent shift toward the Vision-Language and Transformer architectures that characterize modern SOTA baselines. By building upon these established baselines, we demonstrate that rather than maintaining separate experts, a unified model can leverage the commonalities between these architectures to achieve superior performance through joint training and model merging.

\mypara{Model Merging and Weight Editing}
Weight editing has emerged as a computationally efficient paradigm for synthesizing a unified, multi-task model from a collection of single-task specialists. The field is motivated by the ambitious goal of absorbing the capabilities of multiple expert models into a single parameter space without requiring the prohibitive resources of full-scale joint training. Early approaches relied on straightforward Weight Averaging \cite{weightav1,weightav2}. However, these methods often suffered from significant performance degradation due to parameter interference and the lack of a consistent coordinate system across models. This challenge was largely addressed by the introduction of Task Arithmetic and Task Vectors \cite{task_editing}, which define model specialization as a directional vector in weight space relative to a shared initialization. By grounding experts to a common base, task arithmetic allows for the additive combination of capabilities. To further mitigate the interference between competing task vectors, recent advancements have focused on resolving sign conflicts and redundant parameters. Techniques such as TIES-Merging \cite{yadav2023tiesmerging} introduce sign-consensus mechanisms to prevent conflicting weight updates from neutralizing one another. Similarly, sparsification methods like Breadcrumbs \cite{breadcrumbs} isolate the most influential parameters within each task vector, reducing the noise introduced during the fusion process. We build upon these principles in our PSMC framework, utilizing task arithmetic to unify script-specific expertise into a singular, high-performance multilingual backbone.

\section{Datasets and Backbone Architecture}

\myfirstpara{Modality and Language Coverage}
We focus our investigation on word-level recognition for two primary reasons. First, publicly available page-level datasets for Indic scripts remain extremely sparse. Second, the word-level modality offers superior data efficiency; a single digitized page or scene image typically yields hundreds of distinct word-level samples. In extreme data-scarce regimes, this granularity is essential for maximizing the supervisory signal from limited raw data. Our experiments span 10 Indic scripts (supporting 20+ languages) as detailed in Table~\ref{tab:scripts}. Notably, for scripts like Manipuri, which utilize both Meitei and Bengali-Assamese, we include representations for both.

\mypara{Real-World Benchmarks}
We curate our training and evaluation sets from prominent public resources accessible via the NLTM OCR initiative~\cite{ILOCR}. Specifically, we utilize IndicSTR-Roadside~\cite{Indic_Roadside} and IndicSTR12~\cite{IndicSTR12} to represent scene-text environments, and the Mozhi dataset~\cite{Towards_Deployable} for printed-text settings. These datasets constitute the most comprehensive and recent benchmarks for Indic OCR. A detailed statistical summary of these resources is provided in Table~\ref{tab:public_datasets}.

\mypara{Synthetic Data Generation}
To augment our low-resource training, we utilize SynthTiger~\cite{synthtiger} and TRDG~\cite{TRDG} to generate a diverse corpus of synthetic word-level images. Our generation pipeline mimics both scene and printed text by varying typography, chromaticity, background textures, and kerning. In our constrained regime, we generate 240K synthetic samples per language, split equally between scene and printed modalities. We further supplement this with pre-existing synthetic data from IndicSTR12~\cite{IndicSTR12} and IIIT-Synthetic-R~\cite{ILOCR}.

\mypara{Backbone Architecture: Parseq}
We adopt Parseq~\cite{Parseq} as our architectural backbone, as it has consistently demonstrated state-of-the-art performance in both Indic scene and printed text recognition~\cite{IndicSTR12,Indic_Roadside}.
\begin{itemize}
\item Vision Encoder: We utilize a ViT-L architecture comprising 24 self-attention blocks with a hidden dimensionality of 1024.
\item Decoupled Decoder: The decoder consists of 3 Transformer blocks that perform cross-attention with the visual embeddings to generate character-level predictions.
\item Multilingual Charset: To support a unified multilingual model, we define a comprehensive charset of 987 characters spanning all 10 target scripts, enabling the model to learn a shared representation across scripts.
\end{itemize}

\begin{figure}[t]
    \centering
    \includegraphics[width=0.85\linewidth]{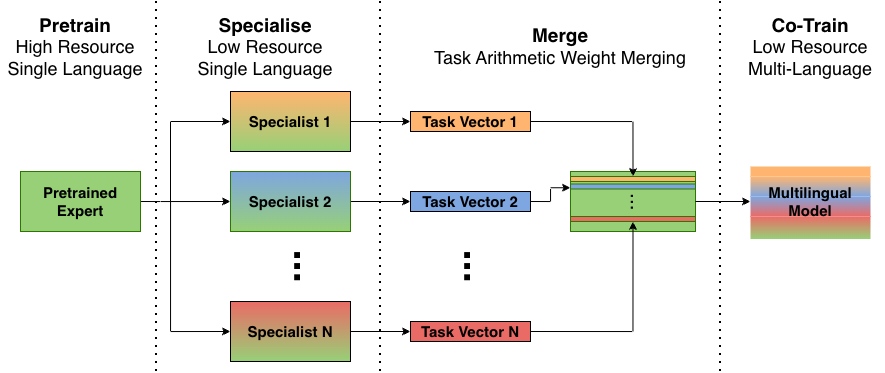}
    \caption{PSMC Framework}
    \label{fig:PSMC}
\end{figure}

\section{Methodology: The PSMC Framework}

We introduce Pre-train, Specialize, Merge, and Co-train (PSMC), a structured fine-tuning paradigm designed to optimize multilingual Indic OCR under extreme low-resource constraints. Our approach is motivated by empirical evidence of cross-script feature alignment in the early stages of visual processing (see Fig. \ref{fig:layer-sim}), allowing the model to explicitly leverage transferable representations across diverse scripts.

\subsection{Transfer Learning from a Pre-trained Anchor}

Transfer learning forms the bedrock of our strategy. Rather than training from a random initialization, which leads to unstable optimization and poor generalization in data-scarce regimes, we initialize our models from a pre-trained anchor model, $\theta_{\text{En}}$. We select English as the anchor language due to the abundance of high-quality, diverse annotated data covering various lighting conditions, domains, and typographies. We train $\theta_{\text{En}}$ on a robust mixture of synthetic and real data, following the original Parseq training protocol \cite{Parseq}. This provides a mature representation space that captures fundamental visual primitives (edges, curves, and strokes) essential for character recognition. As empirically observed in Fig. \ref{fig:layer-sim}, these low-level primitives are largely script-agnostic, facilitating an effective transfer to Indic scripts. For each target language $\ell \in \mathcal{L}$, we derive a specialized model $\theta_{\ell}$ by fine-tuning the anchor model on language-specific data: 
\begin{equation}
\mathcal{L} = \{ \text{Hindi}, \text{Malayalam}, \text{Tamil}, \text{Punjabi}, \dots \}.
\end{equation}
This specialization allows the model to adapt its high-level semantic layers to specific character geometries while retaining the robust low-level features of the anchor.

\begin{figure}[t]
	\centering
	\includegraphics[width=0.75\linewidth]{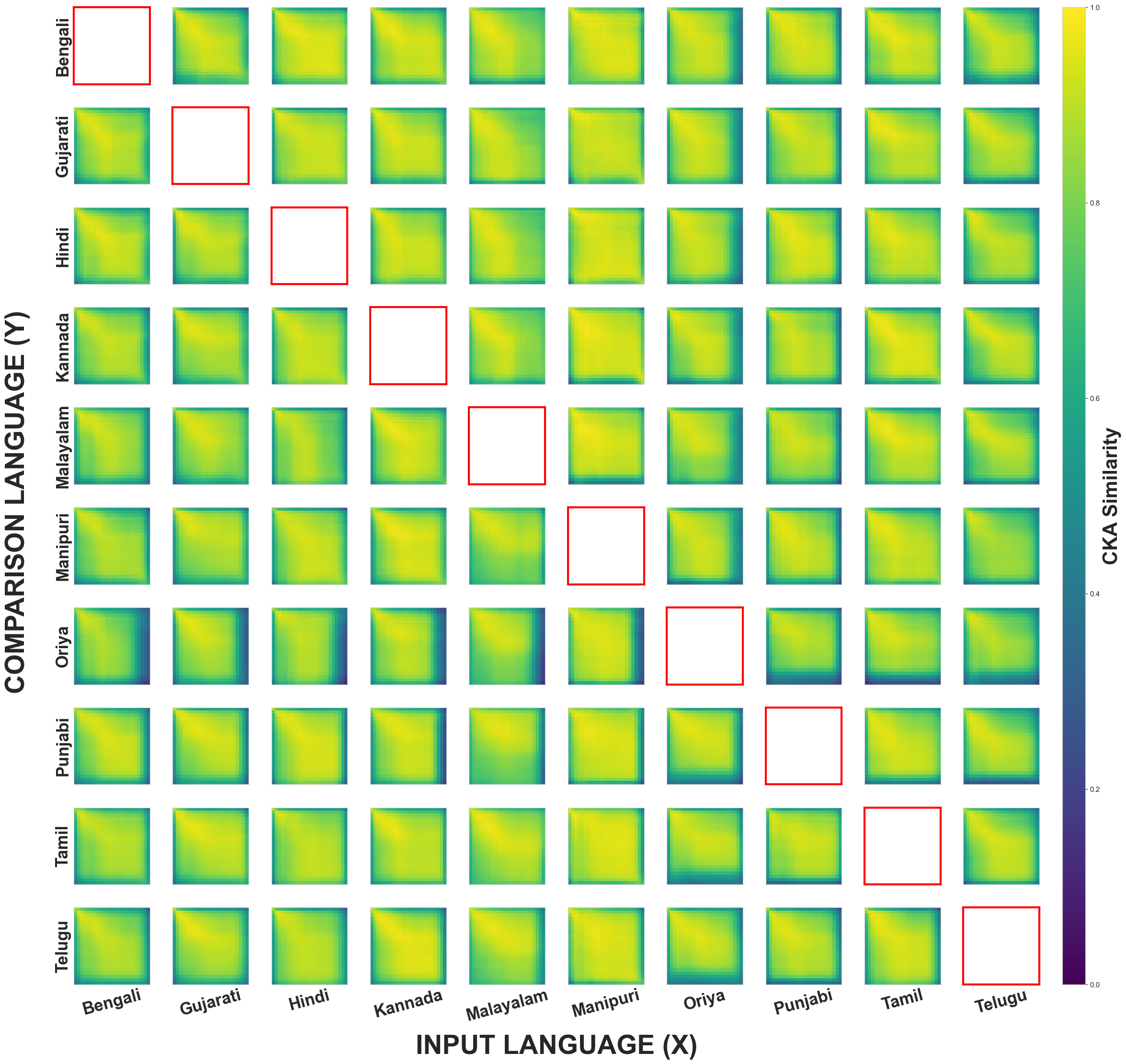}
	\caption{A comprehensive visualization to highlight the feature similarity of layer-wise encoder features across the individual specialist models: Following the same experimental set up as Figure \ref{fig:layer-sim} Heatmap (X,Y) depicts the averaged pairwise CKA feature similarity across all the 24 encoder layers between Model X and Y for input images from language X. Note the high similarity spreads in the upper left corners of all the heatmaps demonstrating the similar high-level, script-agnostic visual cues learnt in the early layers across all specialists.}
	\label{fig:CKA-grid}
\end{figure}

\subsection{Multi-Script Co-Training}

Our analysis of the encoder blocks via Centered Kernel Alignment (CKA) (see Fig. \ref{fig:layer-sim} and Fig. \ref{fig:CKA-grid}) reveals high structural similarity in the initial layers of all expert encoders. This redundancy suggests that separate models for each language independently relearn the same stroke and curve features. To exploit this, we employ multi-script co-training, where the model is jointly optimized across the entire set of target languages. Formally, we construct a joint training distribution over the union of $\mathcal{L}$ to produce a single shared model, $\mathcal{M}_{\text{unified}}$. This exposure to a richer, heterogeneous mixture of character shapes within a single optimization trajectory prompts the network to learn more generalized and robust coarse-level representations. As demonstrated in Tables \ref{tab:combined-mozhi} and \ref{tab:combined-roadside}, this multilingual co-training consistently outperforms isolated fine-tuning under identical per-language data constraints. This confirms that in strictly low-resource regimes, the ``transfer effect'' from multi-script co-training provides measurable gains in both robustness and out-of-distribution generalization.

\begin{figure}[t]
    \centering
    \includegraphics[width=0.8\linewidth]{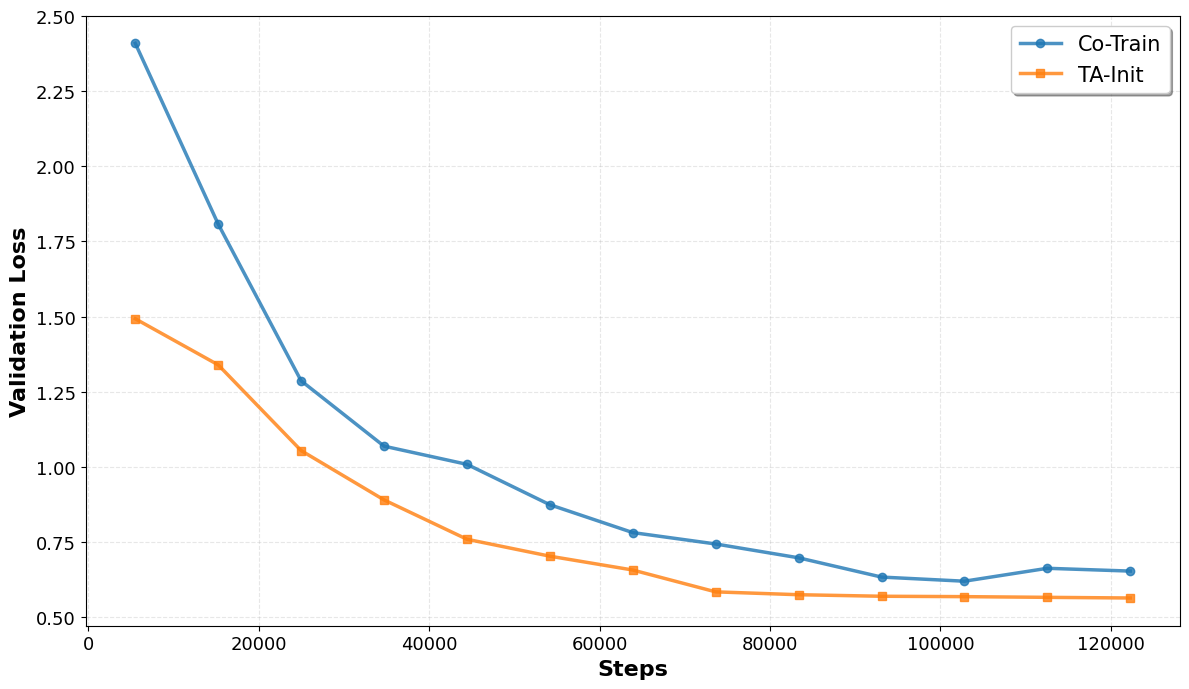}
    \caption{Validation loss for the task arithmetic initialization vs the simple pretrained initialization}
    \label{fig:val_loss}
\end{figure}

\subsection{Task Arithmetic and Joint Refinement}

Weight merging strategies, such as Task Arithmetic, have demonstrated significant success in zero-shot generalization for CLIP-based architectures in image classification tasks \cite{task_editing,breadcrumbs,yadav2023tiesmerging}. While these classification frameworks typically employ dynamically generated task-specific heads, a unified multilingual OCR model requires a singular, shared architecture capable of simultaneous multi-script inference without head-switching. We therefore leverage Task Arithmetic to synthesize an intermediate merged model, $\theta_{\text{Merged}}$. This stage is designed to preserve the shared visual primitives observed in the lower encoder layers while aggregating the divergent, language-specific features captured by individual experts. While $\theta_{\text{Merged}}$ encapsulates latent multilingual knowledge, it often suffers from functional interference, a conflict between weight updates from different scripts. To resolve this, we introduce a final joint refinement stage that aligns the merged weights across the entire architecture. We find that utilizing a merged initialization rather than a simple pretrained anchor leads to faster convergence and superior final accuracy (see Fig. \ref{fig:val_loss}). 

\begin{table}[t]
    \centering
    \begin{tabular}{lcccccccccc}
        \toprule 
        \multirow{2}{*}{\textbf{Language}}& \multicolumn{2}{c}{\textbf{Baseline \cite{Towards_Deployable}}}
        & \multicolumn{2}{c}{\textbf{Skyline}}
        & \multicolumn{2}{c}{\textbf{Ind-FT}}
        & \multicolumn{2}{c}{\textbf{Co-FT}}
        & \multicolumn{2}{c}{\textbf{PSMC}}\\
        \cmidrule{2-11}
         & \textbf{WRR} & \textbf{NED} & \textbf{WRR} & \textbf{NED} & \textbf{WRR} & \textbf{NED} & \textbf{WRR} & \textbf{NED} & \textbf{WRR} & \textbf{NED}\\
        \hline
        Assamese  & 96.5                 & 99                   & 98.38                & 99.6                 & 97.16                & 99.21                & 97.52                & 99.37                & 97.94                & 99.52                \\
        Bengali   & 97.9                 & 99.4                 & 98.45                & 99.55                & 98.27                & 99.49                & 98.55                & 99.63                & 98.71                & 99.64                \\
        Gujarati  & 93.9                 & 96.5                 & 97.12                & 99                   & 94.68                & 98.15                & 96.1                 & 98.61                & 96.29                & 98.72                \\
        Hindi     & 96.3                 & 98.2                 & 98.51                & 99.58                & 97.82                & 99.29                & 98.15                & 99.43                & 98.24                & 99.46                \\
        Kannada   & 90.7                 & 97.7                 & 97.69                & 99.43                & 95.12                & 98.62                & 97.01                & 99.3                 & 97.14                & 99.38                \\
        Malayalam & 97.7                 & 99.7                 & 98.86                & 99.66                & 97.84                & 99.32                & 98.72                & 99.77                & 98.9                 & 99.79                \\
        Manipuri  & 96.9                 & 99                   & 98.14                & 99.33                & 96.19                & 98.74                & 98.05                & 99.34                & 98.19                & 99.32                \\
        Marathi   & 96.9                 & 99.2                 & 96.43                & 97.79                & 95.83                & 98.44                & 98.16                & 99.4                 & 98.06                & 99.38                \\
        Oriya     & 94.8                 & 97.2                 & 95.51                & 98.19                & 93.81                & 97.57                & 95.25                & 98.18                & 95.08                & 98.13                \\
        Punjabi   & 98.7                 & 99.5                 & 99.27                & 99.75                & 97.89                & 99.26                & 99.03                & 99.7                 & 98.82                & 99.55                \\
        Tamil     & 91.8                 & 98                   & 93.32                & 98.45                & 91.22                & 97.58                & 92.67                & 98.33                & 93.06                & 98.39                \\
        Telugu    & 93.6                 & 96.8                 & 98.31                & 99.46                & 96.83                & 99.26                & 98.35                & 99.69                & 98.37                & 99.68                \\
        \midrule
        Average   & 95.48                & 98.35                & 97.5                 & 99.15                & 96.05                & 98.74                & 97.3                 & 99.23                & 97.4                 & 99.25                \\
        \bottomrule
    \end{tabular}
    \vspace{0.5em}
    \caption{WRR and NED scores on the Mozhi Dataset}
    \label{tab:combined-mozhi}
\end{table}

\mypara{Mathematical Formulation}
Following Ilharco et al. \cite{task_editing}, we define a task vector as the displacement in parameter space between a pretrained base model and its fine-tuned counterpart. Let $\theta_{\text{En}}$ denote the parameters of the pretrained English anchor model. For each target language $\ell$, we obtain a specialized expert $\theta_{\ell}$ by fine-tuning $\theta_{\text{En}}$ on language-specific data. The task vector for language $\ell$ is defined as:
\begin{equation}
\tau_{\ell} = \theta_{\ell} - \theta_{\text{En}}
\end{equation}
This vector represents the specific direction in weight space required to adapt the English anchor toward the unique orthographic features of language $\ell$. Given a set of $L$ language experts, we linearly combine their respective task vectors to produce the intermediate merged model:
\begin{equation}
\theta_{\text{Merged}} = \theta_{\text{En}} + \frac{1}{L} \sum_{\ell=1}^{L} \tau_{\ell}.
\end{equation}
By grounding the merging process in $\theta_{\text{En}}$, we ensure that the unified model retains a consistent structural foundation while benefitting from the collective expertise of all specialized language vectors.

\section{Experiments and Results}

\myfirstpara{Evaluation Metrics}
To rigorously evaluate model performance, we employ Normalized Edit Distance (NED), as defined in ICDAR19-ReCTS \cite{ReCTS}. This metric accounts for the relative sequence length and is expressed as:
\begin{equation}
	\text{NED} = 1 - \frac{1}{N} \sum_{i=1}^{N} \frac{D(s_i, \hat{s}_i)}{\max(|s_i|, |\hat{s}_i|)}
\end{equation}
where $N$ is the total number of samples, $D$ is the Levenshtein distance, $s_i$ denotes the ground truth, and $\hat{s}_i$ is the predicted string. We additionally report the Word Recognition Rate (WRR), representing the percentage of word-level predictions that exactly match the ground truth.

\mypara{Data Composition and ``Skyline'' Baselines}
Our ``Skyline'' models represent the upper bound of performance, trained on the full aggregate of available real and synthetic data. The training corpus for each language comprises approximately 80K real printed words (Mozhi), 18.5K real scene-text samples (IndicSTR12, IndicSTR-Roadside), and a massive synthetic set of 16M images (split equally between printed and scene modalities).We establish these baselines using two NVIDIA A100 GPUs with a batch size of 128 and a learning rate of $7 \times 10^{-4}$. As shown in Tables \ref{tab:combined-mozhi} and \ref{tab:combined-roadside}, our individual ``Skyline'' experts consistently outperform existing state-of-the-art baselines across all target scripts.

\begin{table}[t]
    \centering
    \begin{tabular}{ccccccccccc}
        \toprule
        \multirow{2}{*}{\textbf{Language}}
        & \multicolumn{2}{c}{\textbf{Baseline \cite{Indic_Roadside}}}
        & \multicolumn{2}{c}{\textbf{Skyline}}
        & \multicolumn{2}{c}{\textbf{Ind-FT}}
        & \multicolumn{2}{c}{\textbf{Co-FT}}
        & \multicolumn{2}{c}{\textbf{PSMC}}\\
        \cmidrule{2-11}
         & \textbf{WRR} & \textbf{NED} & \textbf{WRR} & \textbf{NED} & \textbf{WRR} & \textbf{NED} & \textbf{WRR} & \textbf{NED} & \textbf{WRR} & \textbf{NED}\\
        \midrule
        Bengali   & 85.34                & 92.75                & 90.9                 & 96.55                & 89.08                & 96.15                & 90.22                & 96.39                & 90.78                & 96.73                \\
        Gujarati  & 81.91                & 88.12                & 84.65                & 94.92                & 80.49                & 93.25                & 81.03                & 93.35                & 82.28                & 93.82                \\
        Hindi     & 87.24                & 95.01                & 88.88                & 95.83                & 87.5                 & 95.75                & 84.26                & 89.31                & 88.92                & 94.59                \\
        Kannada   & 79.27                & 87.64                & 93.18                & 97.76                & 91.92                & 97.5                 & 93.04                & 97.69                & 92.88                & 97.84                \\
        Malayalam & 80.31                & 89.42                & 91.58                & 97.82                & 90.82                & 97.48                & 91.88                & 98.05                & 92.44                & 98.17                \\
        Marathi   & 85.5                 & 94.47                & 87.74                & 96.14                & 88.86                & 96.51                & 85.26                & 90.3                 & 89.8                 & 95.12                \\
        Oriya     & 86.53                & 95.13                & 85.98                & 94.84                & 81.28                & 93.14                & 82.76                & 94.2                 & 83.04                & 94.41                \\
        Punjabi   & 84.27                & 91.46                & 86.58                & 94.46                & 81.8                 & 91.96                & 84.72                & 93.83                & 86.32                & 94.06                \\
        Tamil     & 86.35                & 95.63                & 83.5                 & 95.16                & 79.82                & 94.21                & 81.96                & 94.87                & 82.78                & 95.25                \\
        Telugu    & 84.94                & 92.18                & 85.36                & 94.42                & 84.5                 & 94.35                & 86.02                & 95.25                & 85.42                & 95.02                \\
                \midrule
        Average   & 84.17                & 92.18                & 87.84                & 95.79                & 85.61                & 95.03                & 86.12                & 94.328                & 87.47                & 95.5                 \\
        \bottomrule
    \end{tabular}
    \vspace{0.5em}
    \caption{WRR and NED Scores on the IndicSTR-Roadside Dataset}
    \label{tab:combined-roadside}
\end{table}

\mypara{Representation Similarity Analysis}
To investigate the structural commonalities across specialized experts, we utilize Centered Kernel Alignment (CKA) \cite{CKA}. CKA provides a robust measure of similarity between neural representations that is invariant to orthogonal transformations and isotropic scaling. For two feature matrices $\mathbf{X}$ and $\mathbf{Y}$ extracted from $n$ samples, CKA is computed as:
\begin{equation}
	\text{CKA}(\mathbf{X}, \mathbf{Y}) = \frac{\text{HSIC}(\mathbf{XX}^\top, \mathbf{YY}^\top)}{\sqrt{\text{HSIC}(\mathbf{XX}^\top, \mathbf{XX}^\top) \cdot \text{HSIC}(\mathbf{YY}^\top, \mathbf{YY}^\top)}},
\end{equation}
where HSIC represents the Hilbert-Schmidt Independence Criterion \cite{HSIC}. Our cross-model CKA analysis (Fig. \ref{fig:CKA-grid}) reveals that while deeper layers diverge to capture script-specific nuances, the early layers across all experts remain highly similar. This finding confirms that models independently relearn the same visual primitives, providing a strong empirical justification for our merging and co-training strategy.

\begin{figure}[t]
	\centering
	\begin{subfigure}{0.49\linewidth}
		\centering
		\includegraphics[width=1.0\linewidth]{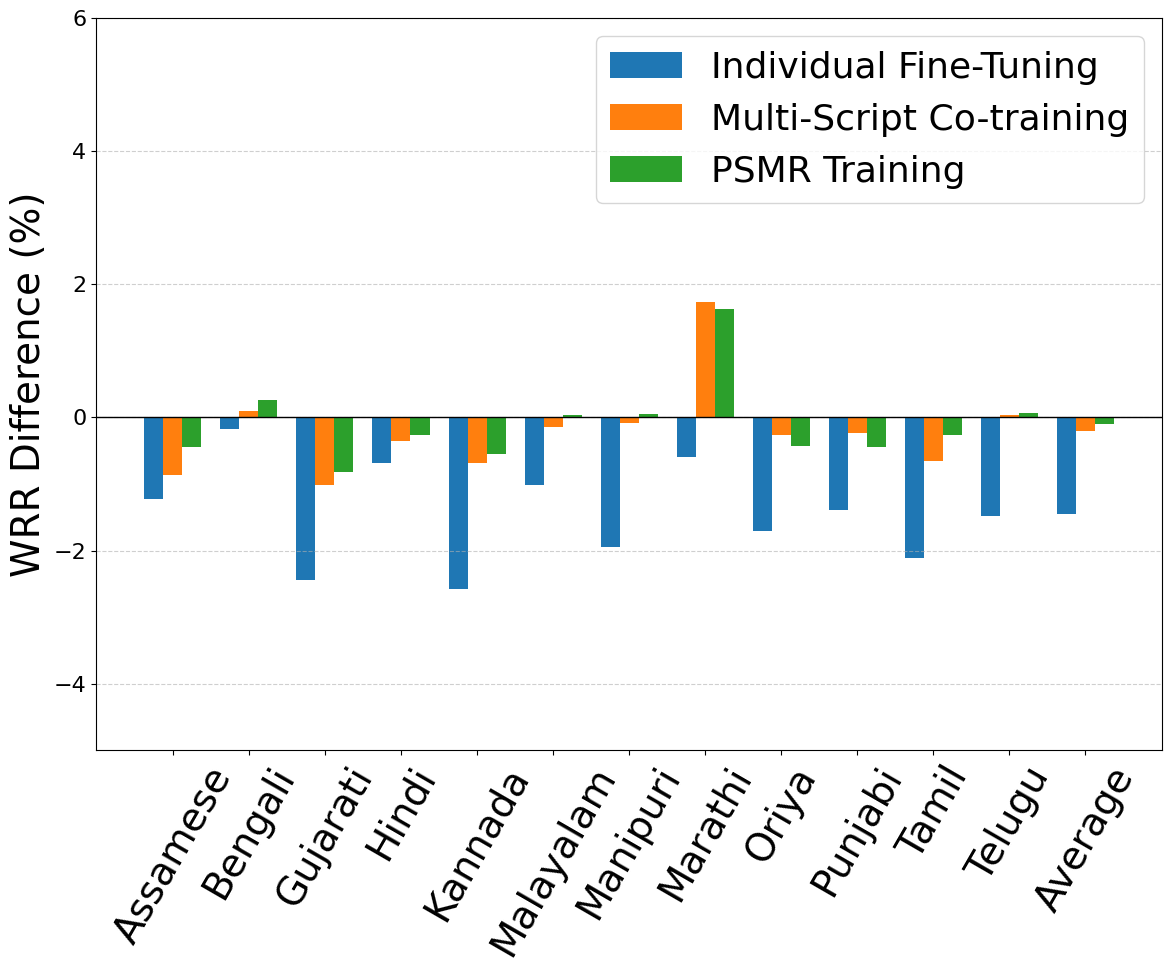}
	\end{subfigure}
	\begin{subfigure}{0.49\linewidth}
		\centering
		\includegraphics[width=1.0\linewidth]{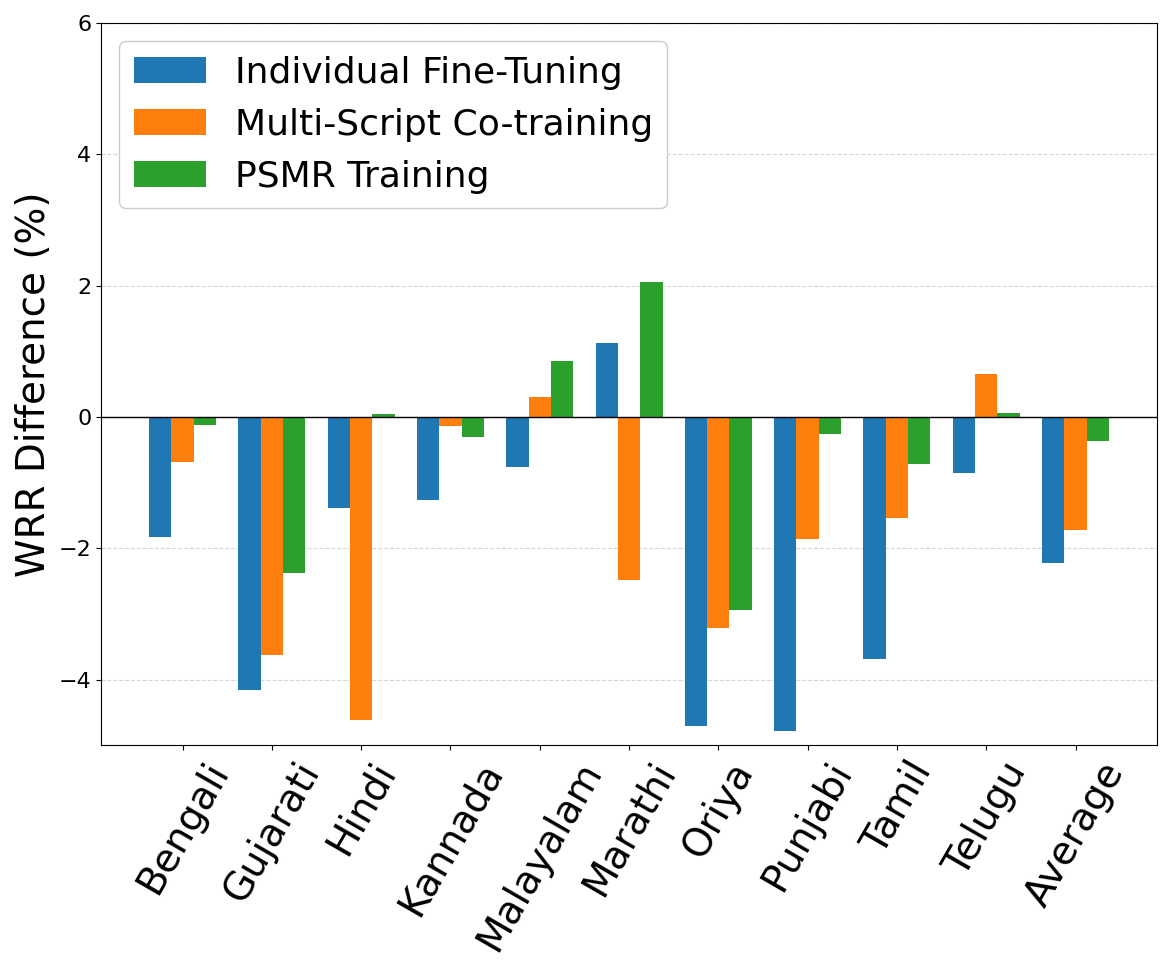}
	\end{subfigure}
	\caption{Drop in WRR with respect to the Skyline WRR metrics for (left) Printed Text, and (right) Scene Text.}
	\label{WRR-Drop}
\end{figure}
\begin{figure}[t]
	\begin{subfigure}{0.5\linewidth}
		\centering
		\includegraphics[width=0.9\linewidth]{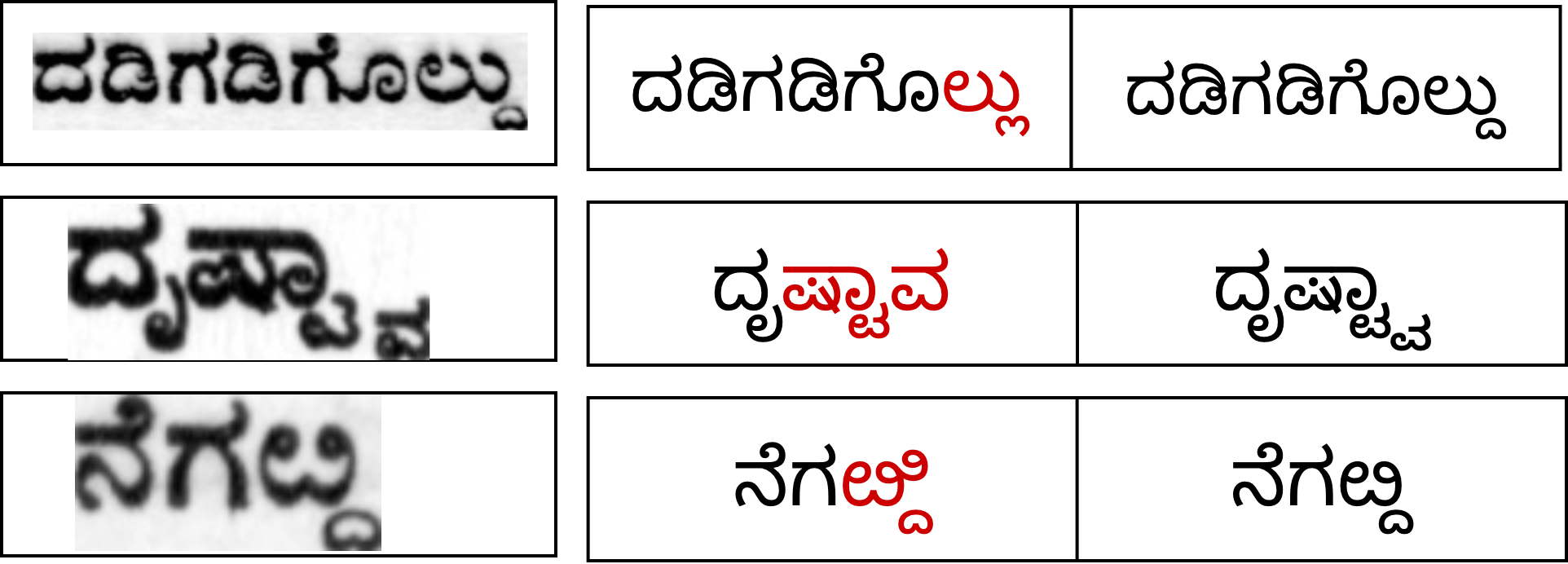}
		\caption{Errors in Kannada vattak-akshara strokes}
	\end{subfigure}
	\begin{subfigure}{0.5\linewidth}
		\centering
		\includegraphics[width=0.8\linewidth]{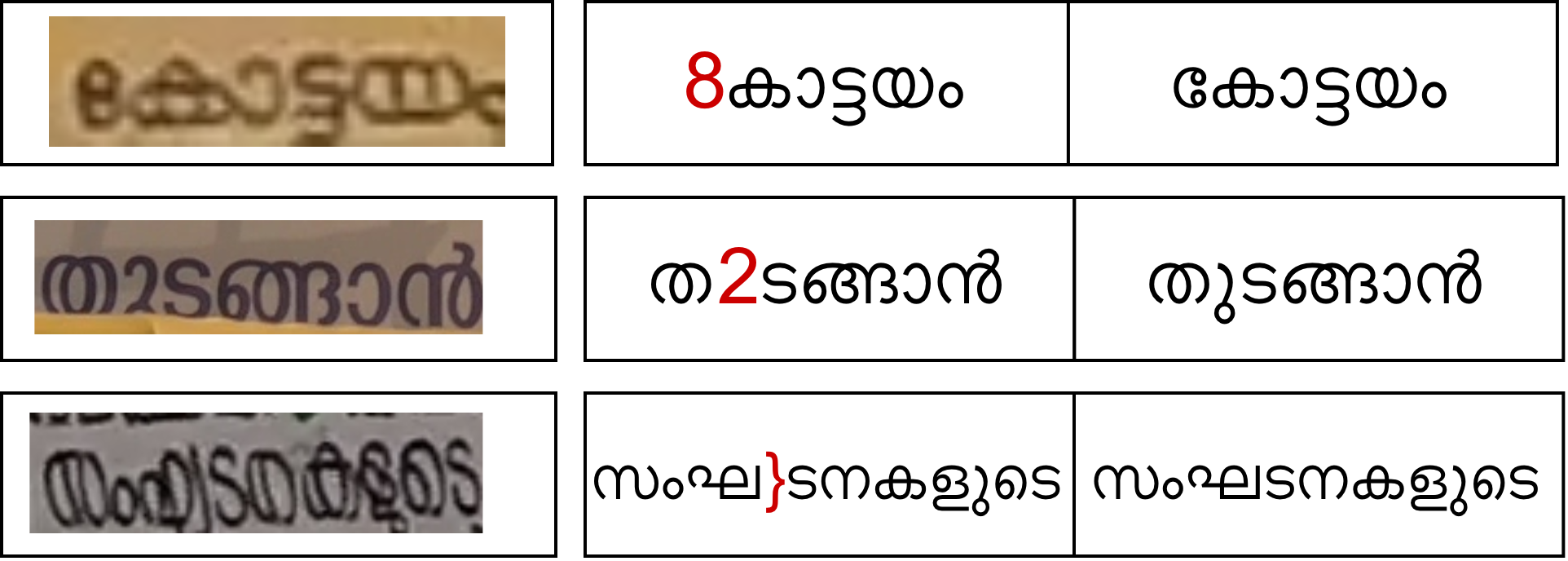}
		\caption{Malayalam letters confused with\\numbers and punctuation}
	\end{subfigure}
	\begin{subfigure}{0.5\linewidth}
		\centering
		\includegraphics[width=0.9\linewidth]{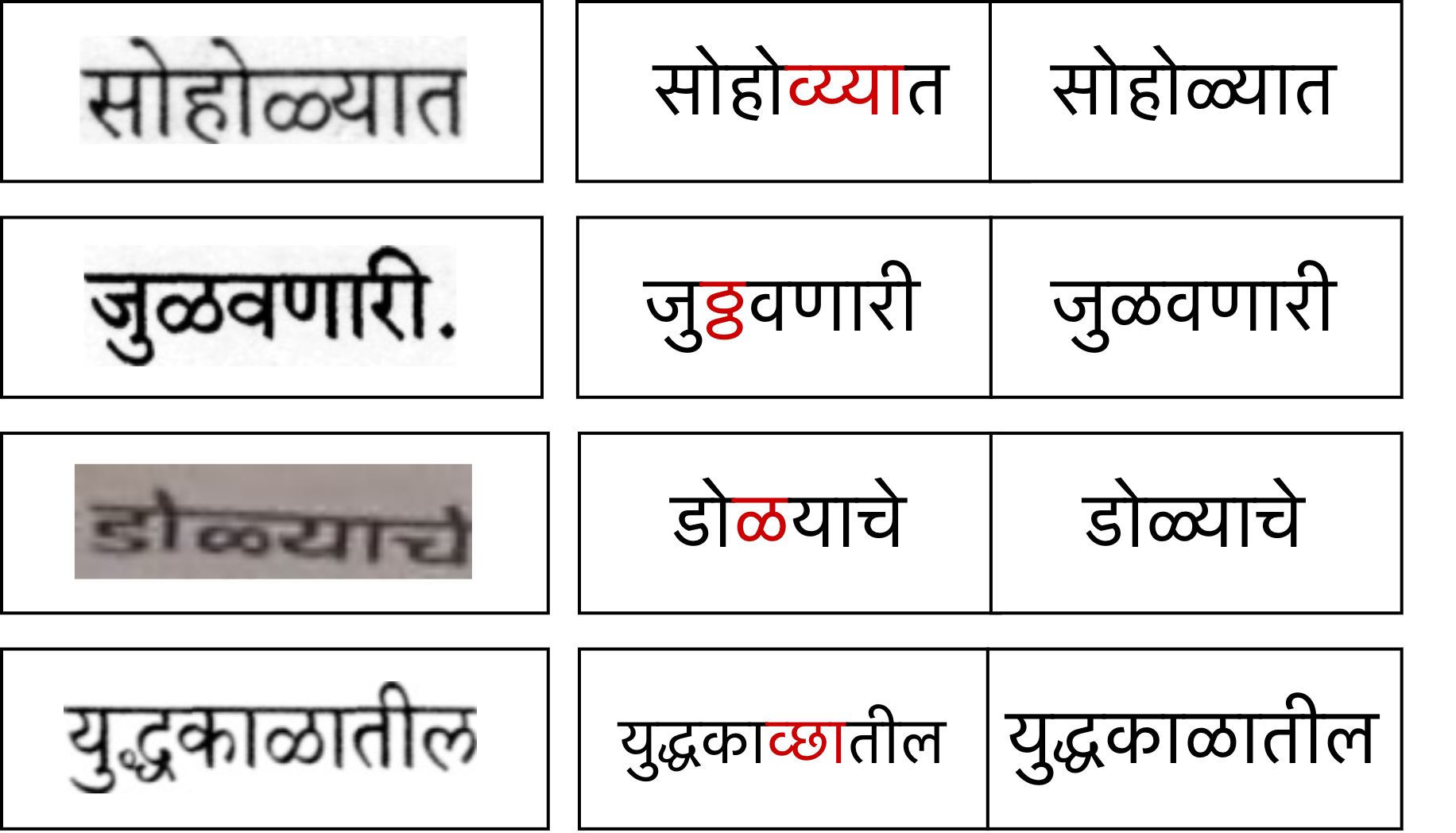}
		\caption{Errors recognising Marathi letter {\MrFont ळ} 
		}
	\end{subfigure}
	\begin{subfigure}{0.5\linewidth}
		\centering
		\includegraphics[width=0.9\linewidth]{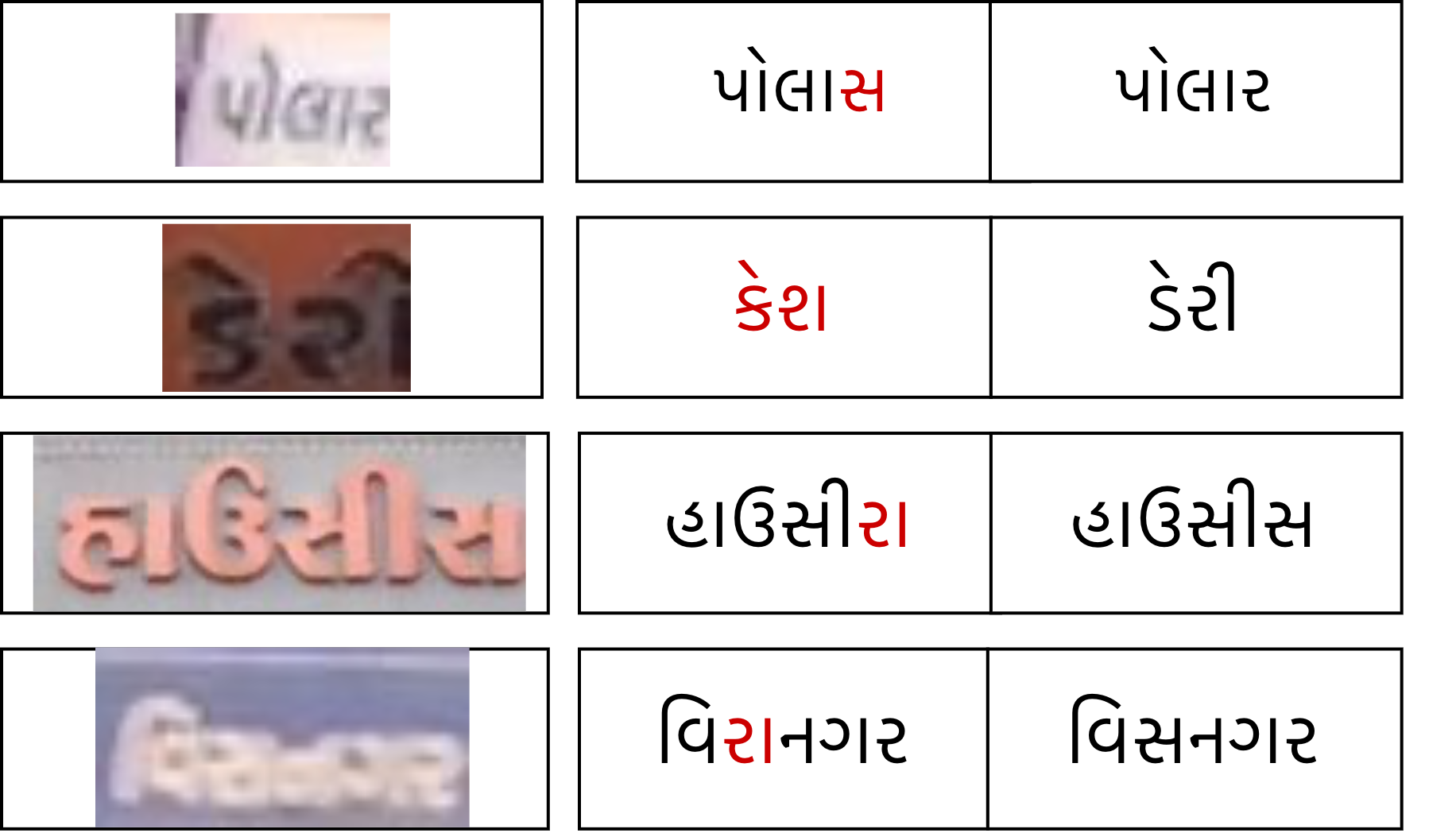}
		\caption{Gujarati letter {\GjFont{ર}} confused with {\GjFont{શ}} and {\GjFont{સ}}
		}
	\end{subfigure}
	\caption{Mistakes in individually finetuned model vs final model}
	\label{fig:script-examples}
\end{figure}

\mypara{Low-Resource Regimes: From Fine-Tuning to PSMC}
In our low-resource experiments, we restrict each language to only 10K real samples (split eqully between scene and printed modalities) and 240K synthetic images.
\begin{itemize}
\item Individual Fine-Tuning: Fine-tuning the English anchor model separately for each script leads to localized specialists but fails to capture cross-script synergies.
\item Naive Co-Training: Jointly training a single model on the union of all low-resource samples yields an average WRR improvement of 1.24\% on Mozhi and 0.51\% on IndicSTR-Roadside compared to individual fine-tuning.
\item PSMC (Task Arithmetic + Co-Training): By initializing the joint training from the merged weight space $\theta_{\text{Merged}}$ (derived via task arithmetic), we achieve the highest performance.
\end{itemize}
As illustrated in Figure \ref{WRR-Drop}, PSMC significantly minimizes the performance gap relative to the ``Skyline'' models, often staying within 0.4\% of the full-data upper bound. Furthermore, the validation loss curves in Fig. \ref{fig:val_loss} demonstrate that PSMC provides a superior initialization, leading to faster convergence and a deeper local minimum compared to naive co-training. Qualitative improvements are detailed in Fig. \ref{fig:script-examples}, showcasing PSMC's ability to resolve subtle character ambiguities that individual fine-tuned models frequently misidentify.

\section{Conclusion}

In this work, we addressed the critical data bottleneck that hinders the extension of robust Vision-Language Models to low-resource Indic scripts. By investigating the structural evolution of specialized OCR experts, we identified a significant hierarchical redundancy: while higher layers diverge to capture script-specific nuances, lower-level visual primitives remain largely script-agnostic. This observation provided the empirical foundation for our PSMC (Pre-train, Specialize, Merge, and Co-train) framework. By utilizing task arithmetic to unify language-specific experts into a singular backbone, followed by joint refinement, we demonstrated that cross-script ``transfer effects'' can compensate for extreme data scarcity. Our results across 10 Indian scripts show that PSMC consistently outperforms individual fine-tuning, achieving an average WRR improvement of $\sim2\%$ and nearing the performance of ``skyline'' models trained on significantly larger datasets. Ultimately, our framework offers a scalable and parameter-efficient pathway for bridging the digital divide in OCR, ensuring that the benefits of modern multimodal AI are accessible to the global long-tail of languages.

\bibliographystyle{splncs04}
\bibliography{mybibliography.bib}

\end{document}